\documentclass[IEEE]{article}
\usepackage[margin=1.25in]{geometry} 

\usepackage{tcolorbox}
\usepackage{graphicx}
\usepackage{biblatex}
\usepackage{graphicx}
\usepackage{xcolor}
\usepackage{soul}
\graphicspath{{figures/}}
\usepackage{subfig}
\usepackage{amsmath}
\usepackage{algorithm}
\usepackage{algpseudocode}

\usepackage[T1]{fontenc}
\usepackage{titling}
\title{\textbf{KG-RAG: Bridging the Gap Between Knowledge and Creativity}}
\author{\textbf{Diego Sanmart\'in} \\
IE University, Spain \\
\texttt{dsanmartin.ieu2020@student.ie.edu}}

\makeatletter
\renewcommand{\@maketitle}{%
  \newpage
  \null
  \vskip 2em%
  \begin{center}%
    \let \footnote \thanks
    {\LARGE \@title \par}%
    \vskip 1.5em%
    {\large
      \lineskip .5em%
      \begin{tabular}[t]{c}%
        \@author
      \end{tabular}\par}%
  \end{center}%
  \par
  \vskip 1.5em}
\makeatother

\begin{document}

\maketitle

\begin{abstract}
Ensuring factual accuracy while maintaining the creative capabilities of Large Language Model Agents (LMAs) poses significant challenges in the development of intelligent agent systems. LMAs face prevalent issues such as information hallucinations, catastrophic forgetting, and limitations in processing long contexts when dealing with knowledge-intensive tasks. This paper introduces a KG-RAG (Knowledge Graph-Retrieval Augmented Generation) pipeline, a novel framework designed to enhance the knowledge capabilities of LMAs by integrating structured Knowledge Graphs (KGs) with the functionalities of LLMs, thereby significantly reducing the reliance on the latent knowledge of LLMs. The KG-RAG pipeline constructs a KG from unstructured text and then performs information retrieval over the newly created graph to perform KGQA (Knowledge Graph Question Answering). The retrieval methodology leverages a novel algorithm called Chain of Explorations (CoE) which benefits from LLMs reasoning to explore nodes and relationships within the KG sequentially. Preliminary experiments on the ComplexWebQuestions dataset demonstrate notable improvements in the reduction of hallucinated content and suggest a promising path toward developing intelligent systems adept at handling knowledge-intensive tasks. 
\footnote{The project website is available at \url{https://dsanmart.github.io/KG-RAG/}}

\end{abstract}

\section{Introduction}

Large Language Models (LLMs) \cite{gpt2} have enabled researchers to achieve significant progress in the development of versatile intelligent agents, emerging as a potential solution for bridging the gap toward achieving Artificial General Intelligence (AGI) \cite{risellmagents}. Notable efforts, such as Langchain \cite{Chase_LangChain_2022} and LlamaIndex \cite{Liu_LlamaIndex_2022} exemplify the community's endeavor to create intelligent LLM-based Agents (LMAs). These LMAs are already being implemented in various real-world applications, such as medicine \cite{tang2023medagents}, and finance \cite{li2023tradinggpt}.

Despite these advancements, the inherent nature of LLMs poses critical challenges that hinder the deployment of LMAs in production environments. One major issue is the disposition of LLMs to generate factually incorrect information, often producing "hallucinated" content, which compromises their reliability \cite{hallucinationsurvey, sirenhallucinationsurvey}. Additionally, these models struggle with processing extended contexts, potentially leading to the loss of relevant information in longer dialogues or documents \cite{lostinthemiddle}. Furthermore, another significant limitation is catastrophic forgetting, where an LLM forgets previously learned knowledge as it is trained with new information \cite{catastrophicforgetting}.

\begin{figure}[h]
\centering
\includegraphics[width=\columnwidth]{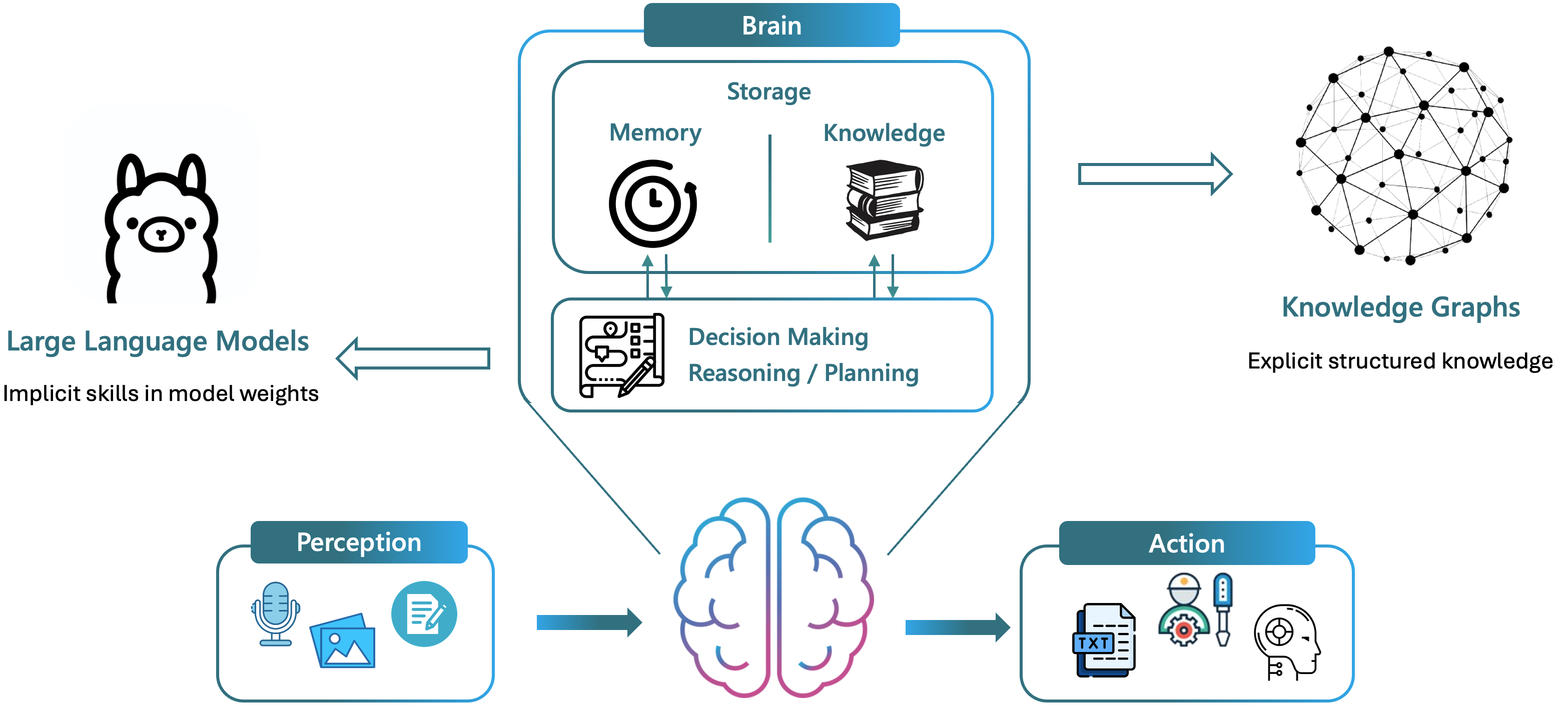}
\caption{shows the three core components of an AI agent: perception, brain, and action. The brain component integrates LLMs for dynamic reasoning and decision-making, alongside KGs for structured knowledge and memory storage.}
\label{fig:agent_workflow}
\end{figure}

As depicted in Figure \ref{fig:agent_workflow}, an AI agent comprises three core components: perception, brain, and action \cite{risellmagents}. These agents are designed to perceive their environments, make decisions, and execute actions, embodying the autonomous operation cycle of Sense-Plan-Act. The diagram illustrates how the brain of an AI agent is the decision-making core that is responsible for reasoning, planning, and storing the agent's knowledge and memories.

From a perspective of human cognition, the analogy of the "extended mind" \cite{clark2010extended}, becomes particularly relevant. Just as humans extend their cognitive reach using tools such as smartphones and notebooks to offload memory and manage complex tasks, LMAs can similarly benefit from external cognitive extensions to enhance their cognitive capacities.


Advancements in multimodal LLMs, which combine inputs from various data types like text, voice, and images \cite{mmllms, li2020oscar}, have significantly enhanced LMAs' perception capabilities, while research enabling interaction with external tools has improved their action capabilities \cite{llmgorilla}. While these aspects continue to evolve, this paper focuses on enhancing the core brain of these agents, fundamentally driven by advancements in prompt engineering and knowledge augmentation.

Prompt engineering techniques like Chain of Thought (CoT) \cite{cotprompt}, Tree of Thought (ToT) \cite{totprompt}, Graph of Thoughts (GoT) \cite{gotprompt}, or ReAct (Reason and Act) \cite{reactprompt} have demonstrated significant improvements in LLMs' reasoning abilities and task-specific actions. Other techniques include Self-Consistency \cite{selfcprompt} and Few-Shot prompting \cite{fewshotprompt}, which have enhanced LLMs' performance and reliability. These techniques have demonstrated that LLMs have promising problem-solving capabilities to take care of the decision-making in the brain of an LMA.

On the other hand, researchers have proposed various techniques for knowledge augmentation on LMAs, including Retrieval-Augmented Generation (RAG), which augments LMAs with external databases and information retrieval tools \cite{ragsurvey, yu2022survey}. RAG dynamically injects specific information into LLMs' prompt at inference time without modifying the model's weights. This method can alleviate hallucinations and outperforms traditional fine-tuning methods by integrating external knowledge more effectively \cite{finetunevsretrieval}, particularly for applications demanding high accuracy and up-to-date information. 

Current RAG implementations rely upon dense vector similarity search as the retrieval mechanism. However, this method which partitions the corpus into text chunks and relies solely on dense retrieval systems, proves inadequate for complex queries \cite{ragllmsurvey}. While some approaches attempt to address this by incorporating metadata filtering or hybrid search techniques \cite{hybridrag}, these methods are constrained by the limited scope of metadata predefined by developers. Furthermore, it remains a difficult challenge to achieve the necessary granularity to find answers to complex queries in similar chunks in the vector space \cite{gao2022precise}. The inefficiency arises from the method's inability to selectively target relevant information, leading to extensive chunk data retrieval that may not contribute directly to answering the queries. An optimal RAG system would accurately retrieve only the necessary content, minimizing the inclusion of irrelevant information. This is where Knowledge Graphs (KGs) can help, by offering a structured and explicit representation of entities and relationships that are more accurate than retrieving information through vector similarity.

KGs enable us to search for "things, not strings" \cite{thingsnotstrings} by storing enormous amounts of explicit facts in the form of accurate, updatable, and interpretable knowledge triples. A triple is essentially a structured representation of a knowledge fact that can be represented as (entity) $-$ [relationship] $\rightarrow$ (entity). Additionally, KGs can actively evolve with continuously added knowledge \cite{nell} and experts can construct domain-specific KGs to provide precise and dependable domain-specific knowledge \cite{abu2021domain}. Prominent KGs such as Freebase \cite{freebase}, Wikidata \cite{wikidata}, and YAGO \cite{yago} exemplify this concept in practice. A significant body of research has emerged at the intersection of graph-based techniques and LLMs \cite{jin2023large, pan2024unifying, li2023survey, pan2023integrating, wang2024can, zhang2023graph}, along with noteworthy applications include reasoning over graphs \cite{luo2023reasoning, saha2021explagraphs, yu2023thought, jiang2023structgpt} and improvements in the integration of graph data and LLMs \cite{chai2023graphllm, yu2023thought}.

In this work, we create a KG-RAG pipeline that extracts triples from raw text, stores them in a KG database, and allows searching complex information to augment LMAs with external KGs that serve as robust and faithful knowledge storage. 
We address the problems of hallucination, catastrophic forgetting, and granularity in dense retrieval systems. We achieve promising results depending on the task, suggesting a promising path toward building LMAs that perform well on knowledge-intensive tasks.


\section{Related Work}

In recent years, integrating Language Model Agents (LMAs) with external databases has significantly advanced knowledge retrieval and memory recall capabilities. Notably, systems like ChatDB \cite{hu2023chatdb} leverage SQL databases for complex reasoning over stored memories, while RET-LLMs [modarressi2023retllm] generate triples from past dialogues to enhance agents with external conversational memory. Our research builds upon these advancements, introducing a novel KG-RAG pipeline that leverages recent developments in KG construction and KGQA, dynamically integrating KGs with LMAs to address issues like hallucination through enhanced granularity in information retrieval.

\textbf{Knowledge Graph Construction}:  Recent progress in this field has focused on bridging unstructured text and structured knowledge representation. Key tasks include Named Entity Recognition (NER) \cite{10184827} and Relationship Extraction (RE) \cite{han-etal-2020-data} to extract the triples forming a KG. Jointly performing these tasks has shown promise in reducing error propagation and improving overall performance \cite{cotype, zheng-etal-2017-joint}. While existing work often focuses on heterogeneous KGs with rigid, domain-specific ontologies \cite{mondal-etal-2021-end, patel2022approach, Dutta2021AMVA, Zhu2023LLMsFK}, we propose a flexible, domain-agnostic, homogeneous KG framework. This adaptability overcomes the limitations of predefined ontologies, accommodating the dynamic nature of knowledge \cite{moon-etal-2019-opendialkg}.

\textbf{Knowledge Graph Question Answering (KGQA)}: KGQA has evolved from rule-based systems to sophisticated architectures handling diverse question types \cite{bordes2015largescale, hu2017ganswering}. Recent LLM integration with KGs, like in Chain-of-Knowledge (CoK) \cite{li2024chainofknowledge} and G-Retriever \cite{he2024gretriever}, further enhances precision and efficiency. Our research leverages these advancements, introducing a unique Chain of Explorations (CoE) method for more precise, contextually relevant lookups within the structured knowledge of KGs.

\textbf{Our Contribution}: The KG-RAG pipeline we propose represents a significant step forward in integrating KGs as external knowledge modules for LMAs. It improves upon existing RAG methods by utilizing dynamically updated KGs, addressing information hallucination through more granular and context-sensitive retrieval processes.
 
\section{Preliminaries}

This section defines the notation and formalizes key concepts related to Language Models (LMs), Knowledge Graphs (KGs), and their integration for the task of Knowledge Graph Question Answering (KGQA).

\textbf{Large Language Models (LLMs)} have presented a new paradigm for task adaptation called "pre-train, prompt, and predict", replacing the traditional “pre-train, fine-tune” procedure. In this approach, general language representations are first learned by the LLM through pre-training. The model is then prompted with a natural language query that specifies the task and the model generates the output directly, without the need for fine-tuning on task-specific labeled data \cite{llmnewtrainparadigm}. Let \(LM\) be a Language Model that takes an input prompt as a sequence of tokens \(x=(x_1,x_2,\ldots,x_q)\) and produces as output a sequence of tokens \(y=(y_1,y_2,\ldots,y_m)\). The model \(LM\) is typically trained to optimize a conditional probability distribution \(P(y|x)\), which assigns a probability to each possible output sequence \(y\) given \(x\). Formally, the probability of the output sequence \(y\) given \(x\) is:
\begin{equation}
P(y|x)=\prod_{i=1}^m P(y_i|y_{<i}, x),
\end{equation}
where \(P(y_i | y_{<i}, x)\) represents the probability of generating token \(y_i\) given \(x\) and the tokens prior in position to \(y_i\), denoted as \(y_{<i}\).

\textbf{Knowledge Graphs (KGs)} serve as structured textual graph representations for real-world entities and interrelations. Heterogeneous KGs typically include multiple types of nodes and relationships, guided by a defined ontology. In contrast, homogeneous KGs consist of only one type of node and relationship. Formally, a homogeneous KG can be defined as \(G = (E, R)\), where \(E\) and \(R\) represent the sets of entities and relationships, respectively. In this context, a triple refers to a basic unit of information represented in the graph, comprising three key components: a subject entity \(e\), a predicate relationship \(r\), and an object entity \(e'\). Formally, a triple $t$ can be represented as \( t=(e, r, e') \). Hence, a KG can be defined as follows:

\begin{equation} \label{kg_def}
G = \big\{t=(e,r,e')\text{, where  } e,e'\in E, r \in R\big\}.
\end{equation}

\textbf{Knowledge Graph Question Answering (KGQA)} aims to respond to questions with information from the KG. For complex questions, this task requires Information Retrieval (IR) of information from multiple triples. When the knowledge is navigable in the graph, we can find the answer through a multi-hop traversal path, represented as \( w_z = e_0 \xrightarrow{r_1} e_1 \xrightarrow{r_2} \ldots \xrightarrow{r_l} e_l \), where \(e_i \in E\) denotes the \(i\)-th entity, \( r_i \) denotes the \( i \)-th relation in the multi-hop path \(z = r_1,r_2,\ldots,r_l\), and \(l\) denotes the length of the path. 
The task of KGQA is given a natural language question \(q\) and a KG \(G\) to design a function \(f\) that predicts an answer \(a \in A_q\) based on knowledge from \(G\). Hence, the predicted answer $a^*$ from multi-hop KGQA can be defined as:

\begin{equation} \label{kgqa_def}
a^* = f(q, G) = \underset{a \in A}{\arg\max} \sum_{z} P(w_z | q, G) \cdot P(a | w_z), 
\end{equation}
where \(A\) denotes the set of possible answers, \(P(w_z | q, G)\) represents the model's capacity to identify the most relevant paths \(w_z\) given the natural language question \(q\) and the knowledge graph \(G\), and \(P(a | w_z)\) is the probability of arriving at the answer \(a\) given the paths \(w_z\).

\subsection{Storage}
The storage process involves converting unstructured text data into a structured KG. This is achieved by the extraction of triples from the text using a language model ($LM$). The $LM$ is trained through a few-shot learning approach \cite{fewshotprompt}, typically involving 5-10 examples.

Given a chunk of text \(T\), the \(LM\) must identify and extract relevant triples represented as \(t = (e,r,e')\), where \(e\) and \(e'\) are entities and \(r\) is the relationship between these entities. The process is modeled as:

\begin{equation}
T \xrightarrow{LM} \{t_i\}_{i=1}^n
\end{equation}

Here, \(LM\) denotes the pre-trained Language Model which processes the text prompt \(T\) that is composed by a natural language instruction, along with several example conversions of text to triples for few-shot learning and the text chunk to extract triples from. The output \(\{t_i\}_{i=1}^n\) consists on the set of extracted triples \(\{(e, r, e')_i\}_{i=1}^n\), where \(i\) indexes each triple from 1 to \(n\).

This setup is designed to optimize the \(LM\) in a task-specific conditional probability distribution \(P(T'|T)\), where \(T'\) is the structured output in the form of triples:

\begin{equation}
P(T' | T) = \prod_{i=1}^n P\big(t_i | T\big)
\end{equation}

\subsection{Retrieval}

The IR process is vital for effectively responding to user queries within a KG. This process entails locating and extracting relevant paths through nodes and relationships within the KG, that lead to the answer sought by the query. Formally, given a natural language query \(q\) and a knowledge graph \(G\), the retrieval task can be modeled as a search problem where the objective is to find paths \(w_z\) within the graph $G$ that connect relevant entities through relationships that are appropriate to the specific question posed by $q$. This is mathematically represented as:

\begin{equation} \label{retrieval_definition}
\text{IR} = \underset{w_z}{\arg \max} P(w_z | q, G).
\end{equation}

This formulation implies that the system is tasked with maximizing the conditional probability $P(w_z | q, G)$, where $w_z$ represents a path in $G$ that is most likely to answer the query $q$.

\subsection{Answer Generation}

In the Answer Generation phase, the LM uses the paths \(w_z\) retrieved from the KG to generate coherent and contextually relevant responses to user queries \(q\). This phase involves synthesizing information from the paths into a format directly addressing the query's requirements. The process is formalized as follows:

\begin{equation}
a = \text{LM}(q, {w_z}).
\end{equation}

To achieve optimal results, the LM is tasked with maximizing the probability of generating the most suitable answer $a$, given the query $q$ and the paths $w_z$. This is represented mathematically as:

\begin{equation}
\underset{a \in A}{\arg \max} P(a | q, {w_z}) = \prod_{j=1}^{m} P(a_j | a_{<j}, q, {w_z}).
\end{equation}

In this equation, $A$ is the set of all possible answers, and $m$ is the length of the generated answer. The term $P(a_j | a_{<j}, q, {w_z})$ reflects the probability of choosing each word $a_j$ in the answer sequence, considering all previously chosen words $a_{<j}$, the query $q$, and the context provided by $w_z$.

This integration employs an LLM configured with a standard RAG prompt as input, which combines the query with the retrieved context and specific instructions to produce accurate answers that pertain to the knowledge in the context given.

\section{Methodology}

Our approach encompasses three primary stages: Storage, Retrieval, and Answer Generation.

\subsection{Storage}

The first stage involves transforming unstructured text data into a structured KG that can be queried later. This process fundamentally relies on the extraction of triples formatted as (entity)$-$[relationship]$\rightarrow$(entity). To facilitate this, we employ a 6-shot learning approach on an LLM, tasking it with extracting as many triples as possible from text chunks, while incorporating examples of text-to-triple conversions in the prompt.

\begin{figure}[h]%
    \centering
    \subfloat[\centering]{{\includegraphics[width=0.4\columnwidth]{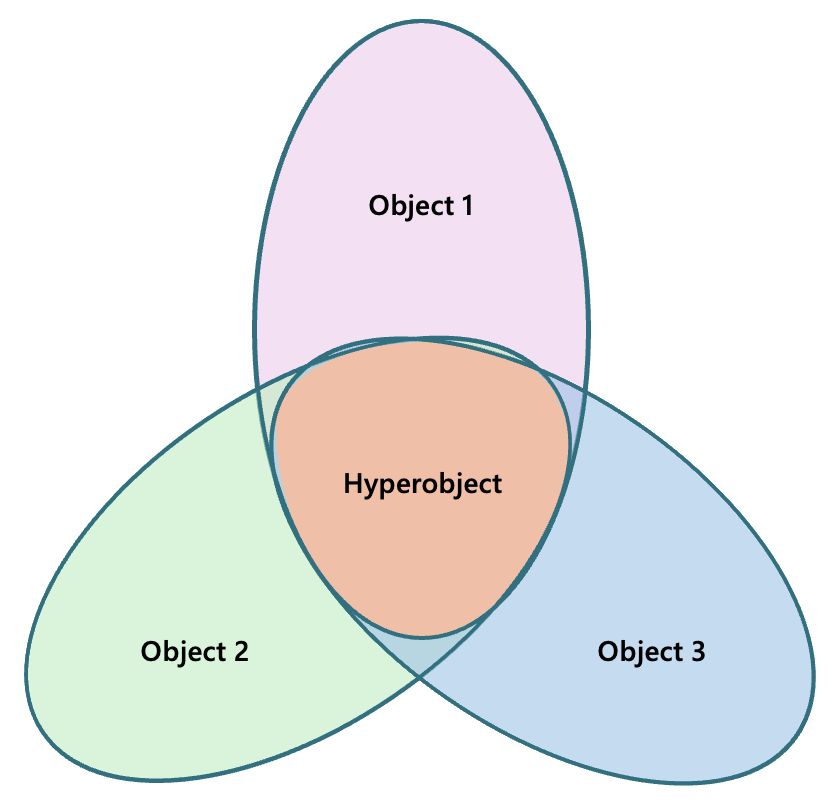} }}%
    \quad
    \subfloat[\centering]{{\includegraphics[width=0.55\columnwidth]{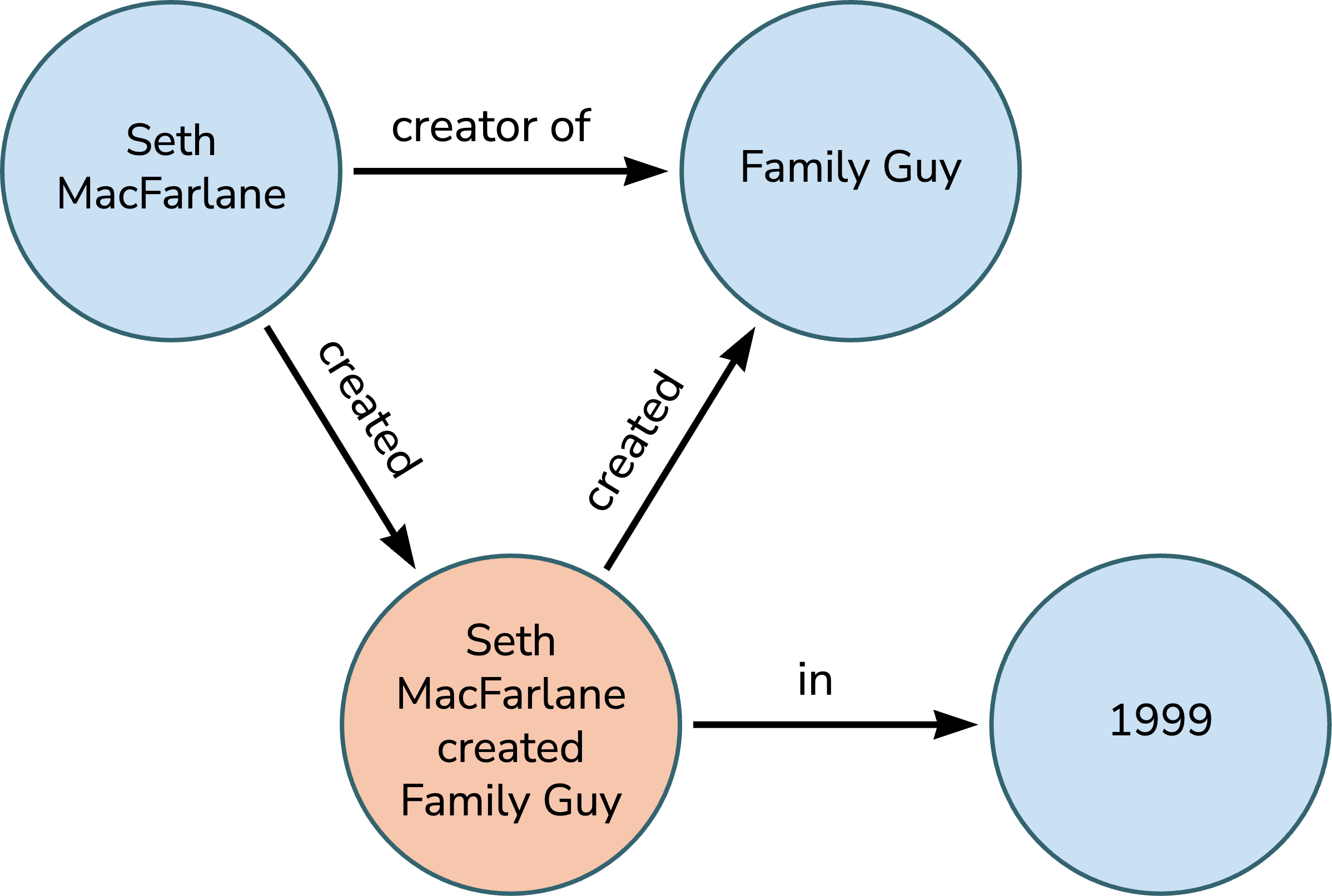} }}%
    \caption{Illustrations of triple hypernode behaviors and representations. \ref{fig:hyperobjects}a) Represents the behavior of hyperobject (triple hypernode) which allows them to connect their internal components to other nodes. \ref{fig:hyperobjects}b) Illustrates our approach to storing triple hypernodes (the orange node) in a traditional KG database.}%
    \label{fig:hyperobjects}%
\end{figure}

To handle more complex informational structures, such as triples linked to specific contexts (e.g. dates), we introduce the concept of \textit{triple hypernodes}. A triple hypernode $h$ can be thought of as a complex node within the set of KG triples $E$ that contains the information of a nested triple. Nested triples represent a relationship between two nodes that may themselves be triple hypernodes. This recursive structure allows for multi-layered relationships within a single node. Additionally, in a triple hypernode, all of the elements that belong to it can be connected to another node by a single relationship. A triple hypernode $h$, represented visually in figure \ref{fig:hyperobjects}a, is defined as:

\begin{equation}
    h^* = (h_1, r_1, h_2), \text{ where } h^*, h_1, h_2 \in E, r_1 \in R.
\end{equation}

Here, $h_1$ and $h_2$ represent subject and target entities respectively that can either be simple entities or other triple hypernodes, allowing $h^*$ to recursively include further nested triple hypernodes:

\begin{equation}
h_i = 
\begin{cases} 
e & \text{if } h_i \text{ is a simple entity} \\
(h_j, r_2, h_k) & \text{if } h_i \text{ is another triple hypernode} ,
\end{cases}
\end{equation}
where $e$ represents a simple entity, $h_j, h_k \in E$, and $r_2 \in R$ denote the relationships within these nested hypernodes.

Equation \ref{eq:hypernode} shows an example of a triple $t$ having several nested layers of triple hypernodes in its entities:

\begin{equation}
\label{eq:hypernode}
t = \big((e_1, r_1, (e_2, r_2, e_3)), r_3, e_4\big),
\end{equation}
where \(e_1\), $e_2$, and \((e_1, r_1, (e_2, r_2, e_3))\) are subject entities,  $r_1$, $r_2$, and $r_3$ are predicate relationships, and $e_3$, $(e_2, r_2, e_3)$ and $e_4$ are object entities. 

Triple hypernodes enable us to store representations of nested and complex relational structures within the KG, enhancing the representation and navigability of the knowledge. For instance, extracting triples from the text "Seth MacFarlane created Family Guy in 1999," would yield the following outputs:

\begin{equation*} \label{hyperedge_example_1}
\begin{aligned}
&\text{(Seth MacFarlane)$-$[is creator of]$\rightarrow$(Family Guy)}\\
&\text{((Seth MacFarlane)$-$[created]$\rightarrow$(Family Guy))$-$[in]$\rightarrow$(1999)},
\end{aligned}
\end{equation*}

where ((Seth MacFarlane)$-$[created]$\rightarrow$(Family Guy)) represents a triple hypernode represented visually in a graph in figure \ref{fig:hyperobjects}b.


To store triple hypernodes in current graph databases, we connect objects to their corresponding hypernodes labeled with the full meaning of the hypernodes and maintain the relationship paths. This allows us to store relationships that connect to that hypernode as shown in figure \ref{fig:hyperobjects}b. Finally, with the KG created, we compute embeddings for all nodes, hypernodes, and relationships and store them in a vector database with their corresponding metadata. This enables us to perform dense vector similarity search through the KG during the retrieval stage of the KG-RAG pipeline.

\subsection{Retrieval}
The retrieval stage constitutes a critical component of the KGQA process, where the goal is to systematically extract pertinent information from a KG in response to a user-specified query. To this end, we perform a Chain of Explorations (CoE) over the KG, a novel approach designed to facilitate the exploration of a KG through a sequential traversal of its nodes and relationships to find relevant information.

\begin{figure}[h]
\centering
\includegraphics[width=0.9\columnwidth]{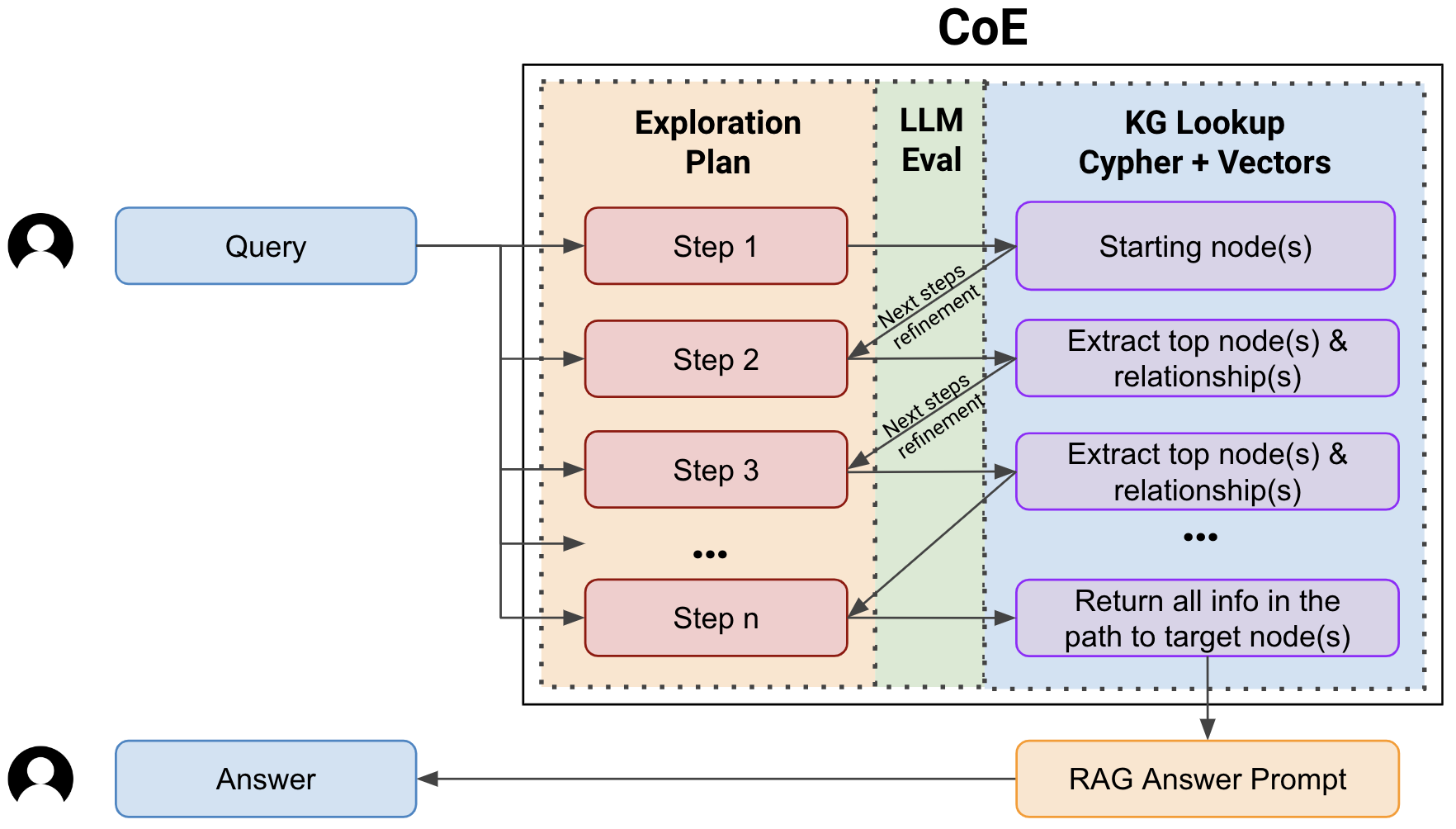}
\caption{Diagram illustrating the components to perform KGQA over a KG using Chain of Explorations (CoE).}
\label{fig:coe}
\end{figure}

CoE, formally defined in Algorithm \ref{algo:coe} and illustrated in Figure \ref{fig:coe}, strategically traverses the KG by navigating through nodes and edges that are directly relevant to the query. It performs a step-by-step exploration to ensure a thorough and directed search, increasing the likelihood of retrieving accurate and relevant information.

The algorithm consists of three main components: planning, KG lookups, and evaluation. The planning phase involves devising a sequence of steps that guide the exploration through the KG. KG lookups are executed using a combination of vector similarity search and cipher queries to identify relevant nodes or relationships. The evaluation phase checks the current state of the traversal against the initial query, deciding whether to continue exploring, refine the exploration strategy, or synthesize an answer based on the gathered context.

\begin{algorithm}[!h]
\caption{Chain of Exploration for KG Retrieval}
\label{algo:coe}
\begin{algorithmic}[1]
\Procedure{CoE}{$q, G$}
    \State $exploration\_plan \gets \text{Plan steps to explore through the graph} $
    \State $current\_nodes \gets \emptyset$
    \State $path\_traveled \gets \emptyset$
    \State $current\_step \gets 0$
    \State $failed\_tries \gets 0$
    \While{$current\_step \leq \text{length}(exploration\_plan) \text{ and } failed\_tries < 3$}
        \If{$\text{is\_node\_exploration}(step)$}
            \State $node\_candidates \gets \text{VectorDB.Search}(step)$
            \State $selected\_nodes \gets \text{LM(prompt, }node\_candidates)$
            \State $path\_traveled.\text{add}(selected\_nodes)$
        \EndIf
        \If{$\text{is\_relationship\_exploration}(step)$}
            \State $explorable\_rels \gets \text{Cypher.Query}(current\_nodes)$
            \State $rel\_candidates \gets 
            \text{VectorSimilarity.Rank}(explorable\_rels, step)$
            \State $selected\_rels \gets \text{LM(prompt, }rel\_candidates)$
            \State $current\_nodes \gets  \text{get\_target\_nodes}(current\_nodes, selected\_rels)$
            \State $path\_traveled.\text{add}(selected\_rels, current\_nodes)$
        \EndIf
        \State $eval\_outcome \gets \text{eval\_state}(path\_traveled, q)$
        \If{$eval\_outcome$ = needs refinement}
            \State $exploration\_plan \gets \text{redefine\_CoT\_steps}(exploration\_plan, q)$
            \State $current\_step \gets 1$
            \State $path\_traveled \gets \{\}$
            \State $failed\_tries \gets failed\_tries + 1$
        \ElsIf{$eval\_outcome$ = continue}
            \State $current\_step \gets current\_step + 1$
        \ElsIf{$eval\_outcome$ = respond}
            \State $\text{return generate\_answer}(path\_traveled, q)$
        \EndIf
    \EndWhile
    \State \textbf{return} "I am sorry, I could not find an answer to your question."
\EndProcedure
\end{algorithmic}
\end{algorithm}

The initial step of CoE uses a few-shot learning prompt along with the user query \(q\) to guide the planner in formulating a strategic exploration plan throughout the KG. In the first step of the plan, CoE finds top-k starting nodes using vector similarity search over a specified keyword and selects relevant nodes to perform further exploration. The process then progresses into a cyclical lookup phase within the KG, outlined as follows:

\begin{enumerate}
    \item Execution of Cypher queries to retrieve connected nodes and relationships in the KG.
    \item Ranking of nodes or relationships by relevance using dense vector embeddings to measure relevance for the current step's task.
    \item Utilizing an LLM to filter and select the most relevant nodes or relationships for continuing the exploration hops over the KG.
\end{enumerate}

This is followed by an evaluation phase, where the LLM assesses the alignment of the current traversal with the initial plan. Based on this assessment, the algorithm decides whether to continue the exploration, adjust the exploratory steps, or synthesize a response based on the context found thus far. Suppose the evaluation phase identifies a conclusive answer within the traversed path. In that case, it compiles the data into a coherent chain of node triples and sends it to the next stage for generating an answer.

\subsection{Answer Generation}
\begin{figure}[ht!]
    \centering
    \includegraphics[width=\columnwidth]{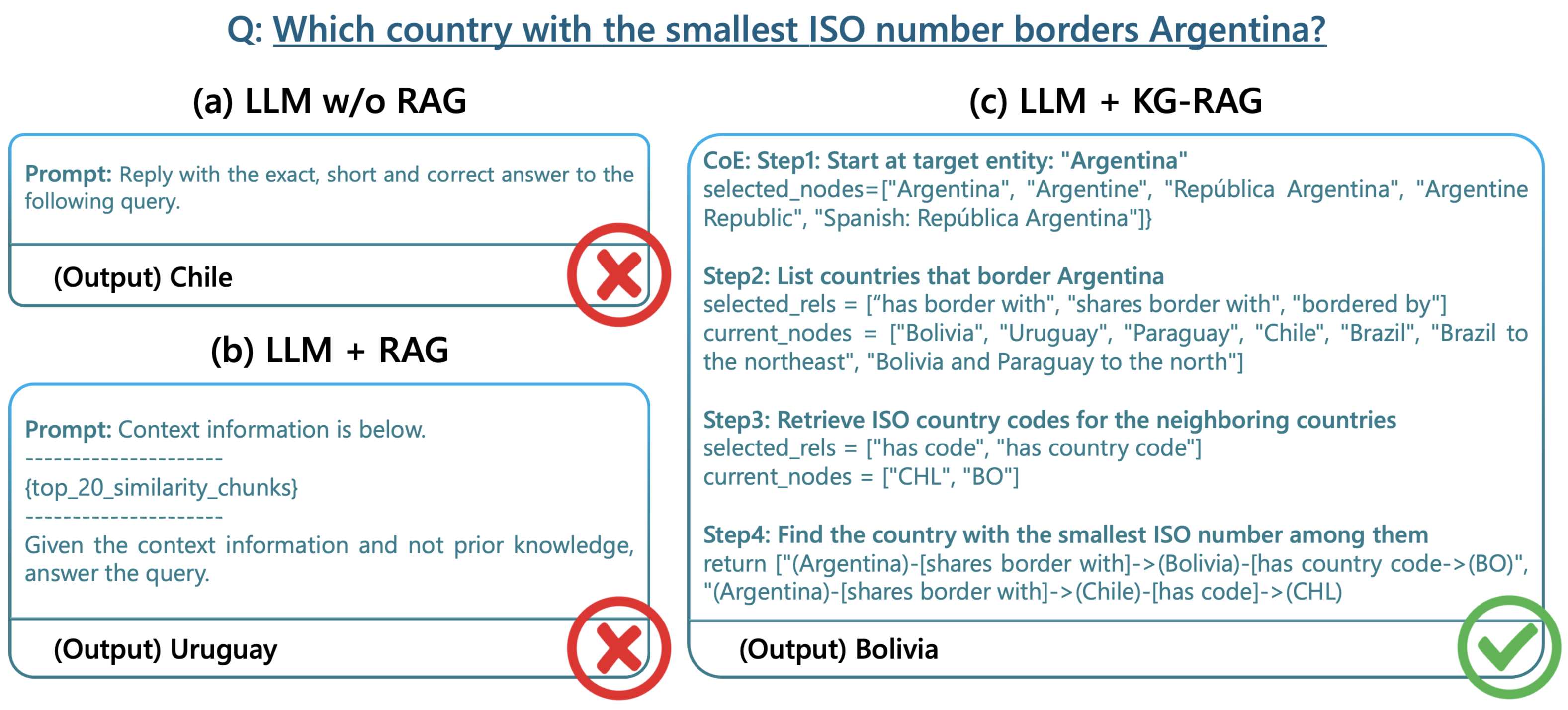}
    \caption{An example comparison of responses to a complex query by (a) LLM without RAG, (b) LLM with vector similarity search RAG, (c) LLM with KG-RAG}
    \label{fig:llm_vs_rag_coe}
\end{figure}

The final stage of the KG-RAG pipeline is generating an answer for the user, where the LLM processes the retrieved KG information to produce accurate and contextually appropriate answers. The LLM is instructed to answer a user question $q$ and only rely on the knowledge found inside the KG at the previous retrieval stage. Figures \ref{fig:coe_example} and \ref{fig:llm_vs_rag_coe} illustrate examples of step-by-step execution of CoE for resolving complex queries.

\begin{figure}[ht!]
\begin{tcolorbox}[width=\textwidth]
\small
\textbf{Q: Which former husband of Elizabeth Taylor died in Chichester?}

\textbf{Step1: Start at target entity: "Elizabeth Taylor"}

selected\_nodes = \text{["Elizabeth Taylor", "Liz Taylor"]}

\textbf{Step2: List former husbands}

selected\_rels = \text{["married to", "married", "was married to"]}

selected\_nodes = \text{["Michael Wilding", "Conrad Nicky Hilton Jr", "Eddie Fisher"]}

\textbf{Step3: Identify places of death for each husband}

selected\_rels = \text{["died in"]}

selected\_nodes = \text{["hospital near his home in Sussex town of Chichester","Chichester,}

\text{West Sussex"]}

\textbf{Step4: Isolate the husband who died in Chichester}

\text{return} \text{(Elizabeth Taylor)-[married]$\rightarrow$(Michael Wilding)-[died in]$\rightarrow$(Chichester,} 

\text{West Sussex)}

\textbf{Answer Generated: Michael Wilding}

\end{tcolorbox}
\caption{An illustrative scenario demonstrating the Chain of Explorations methodology on a complex query involving historical and personal data.}
\label{fig:coe_example}
\end{figure}

\section{Experiments}

In this section, we discuss the evaluation of our approach using the ComplexWebQuestions (CWQ) version 1.1 dataset \cite{talmor18compwebq}. We detail the experiment setup, the dataset used, the metrics chosen to assess performance, and present our results.

\subsection{Experimental Setup}

Our experimental framework is performed on a local machine, using an M2 Ultra with 128 GB of unified memory. We utilize NebulaGraph \cite{wu2022nebula} as our distributed knowledge graph database since it is open-sourced and supports rapid query execution for large-scale graphs, with milliseconds of latency. For vector similarity search, we use a SentenceTransformer \cite{reimers-2019-sentence-bert} model, specifically the \textit{multi-qa-mpnet-base-dot-v1}, for computing high-quality sentence embeddings and store them in a local RedisDB \cite{Redis_2024} instance. Redis serves as our in-memory vector database and with the Redis Search module, we perform fast vector similarity search computations over the DB to find relevant data during the process. The LLM employed is hosted on Azure Cloud and is always GPT-4 Turbo \textit{1106-Preview} configured with a temperature setting of 1, which facilitates the generation of diverse and contextually relevant responses.

\subsection{ComplexWebQuestions Dataset}

The CWQ dataset is specifically designed to challenge systems with complex queries that involve multi-hop reasoning, temporal constraints, and aggregations, making it a challenging task for testing our KGQA framework. The dataset contains 34,689 complex questions, each associated with an average of 367 Google web snippets and the corresponding SPARQL queries for Freebase \cite{freebase}. The questions are composed by merging simple questions into one, such as "What movie starred Miley Cyrus and was released in December of 2003?", to enable multi-hop reasoning over the KG. Additionally, some questions in the dataset are wrongly paraphrased, such as "Who was the leader of France from 1079 until 2012?". This question led CoE to reach the node with the answer "Nicolas Sarkozy". However, the answer generation stage responded by saying it doesn't know the answer since there was no leader of France from 1079 until 2012. The correct question was "Who was the leader of France in 2012, and was the politician that started tenure after 1979?". Other questions, such as "What county is home to Littleton and has CO2 commercial emissions of 0.089006319?", ask about a specific quantity of CO2 emissions that may be in the Freebase KG but not in the Google snippets given to the KG-RAG pipeline.

The dataset was originally split into 80\% for training, 10\% for development, and 10\% for test. For our experiments, we randomly sampled 100 questions from the development split, excluding 11 questions where the correct answers were not present in any of the provided snippets. Since there is an average of 367 web snippets for each question, we filter out the snippets that contain answers to the queries to reduce costs and build a KG of 9604 connected nodes (of which 8141 are normal nodes and 1463 triple hypernodes) and 3175 unique relationship names.

\subsection{Evaluation Metrics}

In our study, we use two standard evaluation metrics for assessing QA systems: Exact Match (EM) and F1 Score \cite{rajpurkar-etal-2016-squad}. Additionally, to address the accuracy of our KG-RAG approach against vector RAG and no RAG specifically, we incorporate a conventional accuracy metric and we introduce a modified version of the precision metric, designed to quantify the incidence of hallucinations. These metrics allow us to assess the model’s ability to provide correct answers and accurately capture the span of the answer within the given context.

\textbf{Exact Match (EM)} which calculates the percentage of predictions that exactly match the ground truth answers. It is defined as:
    \[
    \text{EM }(\%) = \frac{|\{i \mid \hat{a}_i = a_i\}|}{N} \times 100,
    \]
where \( \hat{a}_i \) is the predicted answer, \( a_i \) is the ground truth answer, and \( N \) is the number of samples.

\textbf{F1 Score} considers both the precision and recall of the predicted answers, providing a balance between the two. The F1 score for each instance is calculated as:
    \[
    \text{F1 }(\%) = 2 \times \frac{\text{Precision} \times \text{Recall}}{\text{Precision} + \text{Recall}} \times 100.
    \]
    Precision and recall are computed based on the common tokens between the predicted and the correct answer. It is especially useful in cases where the answers are phrases or sentences.

\textbf{Accuracy} defined as the proportion of answers that have any overlapping word with the ground truth answer:
    \[
    \text{Accuracy }(\%) = \frac{|\{i \mid \text{overlap}(\hat{a}_i, a_i) > 0\}|}{N} \times 100,
    \]
where the overlap function checks if there is any common word between the predicted and the correct answer.

\textbf{Hallucination} quantifies the frequency of "hallucinated" content, defined as responses containing information not present in the ground truth. Hallucinations pose a significant challenge to the accuracy and reliability of QA systems, particularly those utilizing generative models.  The hallucination rate is calculated as follows:
    \[
    \text{Hallucination }(\%) = \frac{|\{i \mid \text{hallucination}(\hat{a}_i, a_i) = 1\}|}{N}  \times 100,
    \]
where $\hat{a}_i$ is the predicted answer, $a_i$ is the ground truth answer, and $N$ is the number of samples. 
A hallucination score of '1' indicates a hallucinated response. This is determined by the absence of perfect precision (i.e., no token mismatch between predicted and ground truth answers) and the presence of specific heuristic indicators, such as phrases like "I don't know," that signal uncertainty and a lack of substantive response. All responses flagged as hallucinations by this metric undergo manual review to ensure the metric's accuracy.

\subsection{Results and Analysis}

We assessed the performance our KG-RAG pipeline on the CWQ dataset, comparing it against a standard configuration for vector similarity search RAG and other state-of-the-art models in the CWQ leaderboard for Google snippets. 

\begin{table}[h]
\begin{tabular}{lcccc}
\hline
\textbf{Model} & \textbf{EM} & \textbf{F1 Score} & \textbf{Accuracy} & \textbf{Hallucination} \\
\hline
Human        & 63                & -                 & -                 & -                 \\
MHQA-GRN        & 33.2                & -                 & -                 & -                 \\
\hline
Embedding-RAG            & 28                & 37                 & 46                 & 30                 \\
KG-RAG         & 19                & 25                 & 32                 & 15                 \\
\hline
\end{tabular}
\centering
\caption{Comparison of model performances against CWQ web snippets dataset}
\label{table:results_comparison}
\end{table}

We present our findings in Table \ref{table:results_comparison} and summarize our analysis, which highlights areas of strength and opportunities for improvement in KG-RAG, below:

\begin{itemize}
    \item KG-RAG achieved an EM of 19\% and an F1 Score of 25\%. These results are lower compared to the top performers on the CWQ leaderboard, where the highest EM by human benchmarks is 63\%, and the best machine performance using the provided Google snippets is 33.2\%. This indicates potential areas for improvement in KG-RAG's ability to accurately parse and retrieve relevant information. Additionally, these results are lower than those of Embedding-RAG, which achieved an EM of 28\% and an F1 Score of 37\%.

    \item KG-RAG demonstrated an accuracy of 32\%, compared to 46\% for Embedding-RAG. This reflects the capability of KG-RAG to retrieve relevant information, but it highlights a challenge in pinpointing the exact answers required for complex multi-hop questions.

    \item Notably, the hallucination rate for KG-RAG stands at 15\%, which is significantly lower than the 30\% observed in Embedding-RAG. This improvement suggests that KG-RAG is more adept at adhering to factual content, reducing the generation of unsupported content significantly compared to Embedding-RAG.
\end{itemize}

The average CoE took between 4 and 5 steps over the KG until reaching the answer nodes. Some questions, such as "What high school did the actor who worked in the movie 'Moment of Truth: Stalking Back' attend?", did not find the starting node ("Moment of Truth: Stalking Back") since web snippets that did not contain the final answer (the name of the high school in this case) were not processed to reduce costs in constructing the KG. This leaves room for improvement in the future work.

\section{Limitations and Future Work}

One major limitation was the financial constraints during our experiments. The high costs associated with converting web snippets to KG triples utilizing LLMs limited our ability to use all of the web snippets corresponding to each question and to perform tests on the entire development split of the CWQ dataset. Future research could expand upon this work as budget allowances increase or as the computing costs of LLMs keep decreasing \cite{shekhar2024towards}.

Additionally, the quality of the CWQ web snippets dataset posed challenges. The information contained within the snippets was often of very low quality and could have affected the reliability of our results. Future iterations of this research may benefit from applying the KG-RAG framework to other question-answer datasets that do not necessarily rely on a traditional knowledge graph structure or in such complicated questions. Furthermore, the performance of the LLM used for KG construction occasionally generated incorrect triples or missed some triple extractions. This error could have propagated through to the performance of CoE. This issue highlights the need for further refinement of the models used for KG construction to ensure accuracy in the triples generated, which are foundational to the subsequent query-answering processes. Future work on integrating advanced methods such as entity resolution \cite{binette2022almost} and linking \cite{shen2021entity} could improve the quality and reliability of the KG used by CoE.

There remains a significant opportunity for future research to develop a specialized dataset for Knowledge Graph Construction using triple hypernodes from raw text. This would allow for fine-tuning of open-source models, such as llama-3-70B \cite{llama3modelcard}, and enable affordable local KG construction from raw text. Such an approach would not only mitigate some of the cost issues noted but also enhance the control and quality of the data used in the research.

\section{Discussion}

The integration of structured knowledge into the operational framework of LMAs through KGs represents a significant paradigm shift in how these agents store and manage their information. The transition from unstructured dense text representations to a dynamic, structured knowledge representation via KGs can significantly reduce the occurrence of hallucinations in LMAs, as it ensures that these agents rely on explicit information rather than generating responses based on knowledge stored "implicitly" in their weights. Additionally, KGs enable LMAs to access vast volumes of accurate and updated information without the need for resource-intensive fine-tuning.

On the technological front, the continuous advancement in hardware optimized for LLM operations, highlighted by innovations like the Groq system \cite{abts2022software}, opens up promising possibilities for improving these integrations. Such advancements significantly speed up information retrieval times, lower operational costs, and boost the efficiency of knowledge retrieval processes. This improvement is crucial for systems like our Chain of Explorations (CoE), where the average response time currently stands at 30 seconds due to multiple calls to the Azure API. With Groq hardware, this could already be reduced to just 3 seconds, making the real-time use of LLMs in conjunction with KGs more practical and effective.

Given these advancements, it's essential that research continues to push the boundaries in KG construction and knowledge representation. Enhancing these areas will further refine how LMAs interact with and utilize structured knowledge, ensuring their reliability for knowledge-intensive tasks. Moreover, research should also aim to empower LLMs to excel in advanced cognitive tasks such as analytical thinking, reasoning, and complex problem-solving. By enabling these agents to 'extend their minds' with external tools, we can facilitate their access to extensive information resources.

\section{Conclusion}

In this paper, we have introduced the KG-RAG pipeline, a novel framework that significantly advances the integration of structured knowledge in Knowledge Graphs (KGs) with the creative and reasoning capabilities of Large Language Models (LLMs) to create advanced Language Model Agents (LMAs). This integration notably reduces the propensity for generating hallucinated content, thereby enhancing the reliability and factual accuracy of responses generated by LMAs. Particularly, our approach to building homogeneous KGs with triple hypernodes sets a new benchmark for deriving structured knowledge from unstructured text and utilizing homogeneous KGs in the grounding of LMAs when answering questions. Within the KG-RAG pipeline, we introduce Chain of Explorations (CoE), a method which sequentially explores nodes and relationships within the KG until it arrives at the information required to answer a question. CoE serves as a testament to how refined methods of knowledge retrieval can leverage the advanced capabilities of LLMs to increase the relevancy and precision of the paths explored through reasoning.

Our preliminary experiments, though limited, show promising results and suggest a robust pathway toward building LMAs that excel in handling knowledge-intensive tasks. This work lays a reliable foundation for further research and development and aims to inspire continued innovations within LMAs, particularly towards developing more advanced LLMs capable of sophisticated cognitive tasks. KG-RAG ensures that the next generation of LMAs not only performs exceptionally across various domains but also adheres to the highest standards of reliability and accuracy.

\printbibliography

\end{document}